\documentclass[11pt,a4paper]{article}
\usepackage{times,latexsym}
\usepackage{url}
\usepackage[T1]{fontenc}

\usepackage{amssymb}
\usepackage{amsmath}
\usepackage{graphicx}
\usepackage{caption}
\usepackage{subcaption}
\usepackage[utf8]{inputenc}

\usepackage{microtype}
\usepackage{enumerate}
\usepackage{tabularx}
\usepackage{arydshln}

\usepackage[acceptedWithA]{tacl2018v2}

\usepackage{xspace,mfirstuc,tabulary}

\newif\iftaclinstructions
\taclinstructionsfalse 
\iftaclinstructions
\renewcommand{\confidential}{}
\renewcommand{\anonsubtext}{(No author info supplied here, for consistency with
TACL-submission anonymization requirements)}
\newcommand{\instr}
\fi

\iftaclpubformat 

\else

\fi

\title{Data-driven Model Generalizability in Crosslinguistic Low-resource Morphological Segmentation}

\author{Zoey Liu \\
Department of Computer Science \\
Boston College \\
 
\texttt{zoey.liu@bc.edu} \\\And
 Emily Prud'hommeaux \\
 Department of Computer Science \\
 Boston College \\
 \texttt{prudhome@bc.edu} \\

}

\date{}

\begin{document}
\maketitle
\begin{abstract}

Common designs of model evaluation typically focus on monolingual settings, where different models are compared according to their performance on a single data set that is assumed to be representative of all possible data for the task at hand. While this may be reasonable for a large data set, this assumption is difficult to maintain in low-resource scenarios, where artifacts of the data collection can yield data sets that are outliers, potentially making conclusions about model performance  coincidental.  
To address these concerns,
we investigate model generalizability in crosslinguistic low-resource scenarios.
Using morphological segmentation as the test case, we compare three broad classes of models with different parameterizations, taking data from 11 languages across 6 language families.
In each experimental setting, we evaluate all models on a \emph{first data set}, then examine their performance consistency when introducing new randomly sampled data sets with the same size and when applying the trained models to unseen test sets of varying sizes.
The results demonstrate that the extent of model generalization depends on the characteristics of the data set, and does not necessarily rely heavily on the data set size. 
Among the characteristics that we studied, the ratio of morpheme overlap and that of the average number of morphemes per word between the training and test sets are the two most prominent factors.
Our findings suggest that future work should adopt random sampling to construct data sets with different sizes in order to make more responsible claims about model evaluation.

\end{abstract}

\section{Introduction}
\label{introduction}

In various natural language processing (NLP) studies, when evaluating or comparing the performance of several models for a specific task, the current common practice tends to examine these models with one particular data set that is deemed representative.
The models are usually trained and evaluated using a predefined split of training/test tests (with or without a development set), or using cross-validation.
Given the results from the test set, 
authors typically conclude that certain models are better than others~\citep{devlin-etal-2019-bert}, report that one model is generally suitable for particular tasks~\citep{rajpurkar-etal-2016-squad,wang-etal-2018-glue,kondratyuk-straka-2019-75}, or state that large pretrained models have ``mastered" linguistic knowledge at human levels~\citep{hu-etal-2020-systematic}. 

Consequently, the ``best" models, while not always surpassing other models by a large margin, are subsequently cited as the new baseline or adopted for state-of-the-art comparisons in follow-up experiments. These new experiments might not repeat the comparisons that were carried out in the work that identified the best model(s).
This model might be extended to similar tasks in different domains or languages, regardless of whether these new tasks have characteristics comparable to those where the model was demonstrated to be effective.

That being said, just because a self-proclaimed authentic 
Chinese restaurant is very good at cooking orange chicken, that does not mean its orange chicken will be delicious on every visit to the restaurant.
Similarly, common implementations of model comparisons have been called into question. 
Recent work by~\citet{gorman-bedrick-2019-need} (see also~\citet{szymanski-gorman-2020-best} and~\citet{sogaard-etal-2021-need}) analyzed a set of the best part-of-speech (POS) taggers using the the Wall Street Journal (WSJ) from the Penn Treebank~\citep{marcus1993building}. 
They demonstrated inconsistent  model performance with randomly generated splits of the WSJ data in contrast to the predefined splits as used in previous experiments~\citep{collins-2002-discriminative}.

Others (see~\citet{linzen-2020-accelerate} for a review) have argued that the majority of model evaluations are conducted with data drawn from the same distribution -- that is, the training and the test sets bear considerable resemblance in terms of their statistical characteristics -- and therefore these models will not generalize well to new data sets from different domains or languages~\citep{mccoy-etal-2019-right,mccoy-etal-2020-berts,jia-liang-2017-adversarial}.
In other words, to evaluate the authenticity of a Chinese restaurant, one would need to not only try the orange chicken, but also order other traditional dishes.

Previous work on model evaluation, however, has focused on high-resource scenarios, in which there is abundant training data. 
Very little effort has been devoted to exploring whether the performance of the same model will generalize in low-resource scenarios~\citep{hedderich2020survey,ramponi-plank-2020-neural}.
Concerns about model evaluation and generalizability are likely to be even more fraught when there is a limited amount of data available for training and testing.

Imagine a task in a specific domain or language that typically utilizes just one specific data set, which is common in low-resource cases.
We will refer to the data set as the \textbf{first data set}.
When this first data set is of a reasonable size, it may well be the case (though it might not be true) that the data is representative of the population for the domain or language, and that the training and the test sets are drawn more or less from the same distribution.
The performance of individual models and the rankings of different models can be expected to hold, at the very least, for new data from the same domain.

Now unfold that supposition to a task in a specific domain or language situated in a low-resource scenario.
Say this task also resorts to just one particular data set, which we again call the first data set.
Considering the limitation that the data set size is relatively small, 
one might be skeptical of the claim that this first data set could represent
the population distribution of the domain or language.
Accordingly, within the low-resource scenario, model evaluations obtained from just the first data set \textit{or any one data set} could well be coincidental or random.
These results could be specific to the characteristics of the data set, which might not necessarily reflect the characteristics of the domain or language more broadly given the small size of the data.
This means that even within the same domain or language, the same model might not perform well when applied to additional data sets of the same sizes beyond the first data set, and the best models based on the first data set might not outperform other candidates when facing new data.
Continuing the Chinese restaurant analogy, the same orange chicken recipe might not turn out well with a new shipment of oranges.

As an empirical illustration of our concerns raised above, this paper studies model generalizability in crosslinguistic low-resource scenarios.
For our test case, we use morphological segmentation, the task of decomposing a given word into individual morphemes (e.g., \emph{modeling} $\rightarrow$ \emph{model + ing}). 
In particular, we focus on surface segmentation, where the concatenation of the segmented sub-strings stays true to the orthography of the word~\citep{cotterell-etal-2016-joint}.\footnote{This includes segmentation for both inflectional and derivational morphology; we used the term ``sub-strings", since not all elements in the segmented forms will be a linguistically-defined morpheme due to the word formation processes. Therefore surface segmentation is in opposition to canonical segmentation; see~\citet{cotterell-etal-2016-joint} for detailed descriptions of the differences between the two tasks.}   
This also involves words that stand alone as free morphemes (e.g., \emph{free} $\rightarrow$ \emph{free}).

Leveraging data from 11 languages across six language families, we compare the performance consistency of three broad classes of models (with varying parameterizations) that have gained in popularity in the literature of morphological segmentation.
In each of the low-resource experimental settings (Section~\ref{evaluate1}), we ask:
\begin{enumerate}[(1)]
    \item To what extent do individual models generalize to new randomly sampled data sets of the same size as the first data set?
    \item For the model that achieves the best results overall across data sets, how does its performance vary when applied to new test sets of different sizes?

\end{enumerate}

\section{Prior Work: Model Generalizability}


A number of studies have noted inconsistent performance of the same models across a range of generalization and evaluation tasks.
For example, ~\citet{mccoy-etal-2020-berts} demonstrated that the same pretrained language models, when paired with classifiers that have different initial weights or are fine-tuned with different random seeds, can lead to drastically unstable results~\citep{mccoy-etal-2019-right}.
This pattern seems to hold for both synthetic~\citep{weber-etal-2018-fine,mccoy2018} and naturalistic language data~\citep{zhou-etal-2020-curse,reimers-gurevych-2017-reporting}.
Using approximately 15,000 artificial training examples, ~\citet{DBLP:conf/icml/LakeB18} showed that sequence-to-sequence models do not generalize well with small data sets (see also~\citet{akyurek-andreas-2021-lexicon}).
Through a series of case studies,~\citet{dodge-etal-2019-show} demonstrated that model performance showed a different picture depending on the amount of computation associated with each model (e.g., number of iterations, computing power).
They suggested that providing detailed documentation of results on held-out data is necessary every time changes in the hyperparameters are applied.

Another line of research has focused on testing the generalizability of large language models with controlled psycholinguistic stimuli such as subject-verb agreement~\citep{linzen-etal-2016-assessing}, filler-gap dependencies~\citep{wilcox-etal-2019-structural} and garden-path sentences~\citep{futrell-etal-2019-neural}, where the stimuli data share statistical similarities with the training data of the language models to different degrees~\citep{wilcox-etal-2020-structural,hu-etal-2020-systematic,thrush-etal-2020-investigating}.
This targeted evaluation paradigm~\citep{marvin-linzen-2018-targeted,doi:10.1146/annurev-linguistics-032020-051035} tends to compare model performance to behaviors of language users.
It has been claimed that this approach is able to probe the linguistic capabilities of neural models, as well as shed light on humans' ability to perform grammatical generalizations.
Thus far, related studies have mainly investigated English, with notable exceptions of crosslinguistic extensions to other typologically diverse languages, such as Hebrew~\citep{mueller-etal-2020-cross}, Finnish~\citep{dhar-bisazza-2021-understanding}, and Mandarin Chinese~\citep{wang-etal-2021-controlled,xiang-etal-2021-climp}.




As fruitful as the prior studies are, 
they are mostly targeted towards evaluating models that have access to large or at least reasonable amount of training data. 
In contrast, there is, as yet, no concrete evidence regarding whether and to what extent models generalize when facing limited data. 

Additionally, although recent work has proposed different significance testing methods in order to add statistical rigor to model evaluations~\citep{gorman-bedrick-2019-need,szymanski-gorman-2020-best,dror-etal-2018-hitchhikers},
most tests make assumptions on the distributional properties or the sampling process of the data (e.g., data points are sampled independently~\citep{sogaard-2013-estimating}), and a number of these tests such as bootstrapping are not suitable when data set size is extremely small due to lack of power (see also~\citet{card-etal-2020-little}).

\section{Morphological Surface Segmentation}

Why morphological surface segmentation?
First, previous work has demonstrated that morphological supervision is useful for a variety of NLP tasks, including but not limited to  machine translation~\citep{clifton-sarkar-2011-combining}, dependency parsing~\citep{seeker-cetinoglu-2015-graph}, bilingual word alignment~\citep{eyigoz-etal-2013-simultaneous}, and language modeling~\citep{blevins-zettlemoyer-2019-better}.
For languages with minimal training resources, surface or subword-based segmentation is able to effectively mitigate data sparsity issues~\citep{tachbelie2014,ablimit}.
For truly low-resource cases, such as endangered and indigenous languages, morpheme-level knowledge has the potential to advance development of language technologies such as automatic speech recognition~\citep{afify2006use} in order to facilitate community language documentation.

Second, information about morphological structures is promising for language learning. 
Especially for indigenous groups, prior studies have proposed incorporating morphological annotation into the creation of online dictionaries as well as preparing teaching materials, with the goal of helping the community's language immersion programs~\citep{garrett2011online,spence2013language}.

Third, despite its utility in different tasks, 
semi-linguistically informed subword units are not as widely used as they might be, since
acquiring labeled data for morphological segmentation in general, including the case of surface segmentation, requires extensive linguistic knowledge from native or advanced speakers of the  language.
Linguistic expertise can be much more difficult to find for critically endangered languages, which, by definition~\citep{meek2012we}, are spoken by very few people, many of whom are not native speakers of the language.
These aforementioned limitations necessitate better understanding of the performance for different segmentation models and how to reliably estimate their effectiveness.

\section{Meet the Languages}
\label{meetlanguage}

A total of eleven languages from six language families were invited to participate in our experiments. 
Following recently proposed scientific practices for computational linguistics and NLP~\citep{bender-friedman-2018-data,gebru2018datasheets}, we would like to  introduce these languages and their data sets explored in our study (Table~\ref{tab:languages}).


First we have three indigenous languages from the Yuto-Aztecan language family~\citep{baker1997complex}, including Yorem Nokki (Southern dialect)~\citep{freeze} spoken in the Mexican states of Sinaloa and Sonora, Nahuatl (Oriental branch)~\citep{lastra} spoken in Northern Puebla, and Wixarika~\citep{gomez} spoken in Central West Mexico.
These three Mexican languages are highly polysynthetic with agglutinative morphology.
The data for these languages was originally digitized from the book collections of \textit{Archive of Indigenous Language}. 
Data collections were carried out and made publicly available by the authors of~\citet{kann-etal-2018-fortification} given the narratives in their work.

Next are four members from the Indo-European language family: English, German, Persian, and Russian, all with fusional morphological characteristics. The data for English was provided by the 2010 Morpho Challenge~\citep{kurimo-etal-2010-morpho}, a shared task targeted towards unsupervised learning of morphological segmentation.
The German data came from the CELEX lexical database~\citep{baayen1996celex} and was made available by~\citet{cotterell-etal-2015-labeled}.
The Persian data from~\citet{persiandata} (see also~\citet{ansari-etal-2019-supervised}) contains crowd-sourced annotations, while the data for Russian was extracted from an online dictionary~\citep{sorokin2018deep}.

The remaining languages each serve as a representative of a different language family.
The data sets for Turkish from the Turkic language family and Finnish from the Uralic language family were again provided by the 2010 Morpho Challenge~\citep{kurimo-etal-2010-morpho}.
The data collection for both Zulu and Indonesian was carried out by~\citet{cotterell-etal-2015-labeled}.
The Zulu words were taken from the Ukwabelana Corpus~\citep{spiegler-etal-2010-ukwabelana}, and the segmented Indonesian words were derived via application of a rule-based morphological analyzer to an Indonesian-English bilingual corpus.\footnote{\url{https://github.com/desmond86/Indonesian-English-Bilingual-Corpus}}
All four of these languages have agglutinative morphology.

\begin{table*}[h!]
\scriptsize
    \centering
    \begin{tabular}{c|c|c|c|c|c}
    \hline
     \textbf{Language} & \textbf{Language family} & \textbf{Morphological feature(s)}  &\textbf{$N$ of  types} &  \textbf{Data set sizes} & \textbf{New test set sizes} \\
     & & & \textbf{in initial data} & & \\ \hline
     & & & & & \\
      Yorem  Nokki  & Yuto-Aztecan & Polysynthetic  & 1,050 & \{500, 1,000\} & --- \\
       & & & & & \\
      Nahuatl &  & Polysynthetic  & 1,096 & \{500, 1,000\} & --- \\
       & & & & & \\
      Wixarika &  & Polysynthetic & 1,350 & \{500, 1,000\} & \{50, 100\} \\ 
      & & & & & \\
      English & Indo-European & Fusional & 1,686 & \{500, 1,000, 1,500\} & \{50, 100\} \\
       & & & & & \\
      German &  & Fusional & 1,751 & \{500, 1,000, 1,500\} & \{50, 100\} \\ 
       & & & & & \\
      Persian & & Fusional & 32,292 & \{500, 1,000, 1,500 & \{50, 100, 500, 1,000\} \\
      & & & & 2,000, 3,000, 4,000\} & \\
       & & & & & \\
      Russian &  & Fusional & 95,922 & \{500, 1,000, 1,500\} & \{50, 100, 500, 1,000\} \\
       & & & & 2,000, 3,000, 4,000\} & \\
       & & & & & \\
      Turkish & Turkic & Agglutinative & 1,760 & \{500, 1,000, 1,500\} & \{50, 100\} \\
      & & & & \\
      Finnish & Uralic & Agglutinative & 1,835 & \{500, 1,000, 1,500\} & \{50, 100\}\\
      & & & & & \\
      Zulu & Niger-Congo & Agglutinative & 10,040 & \{500, 1,000, 1,500\}  & \{50, 100, 500, 1,000\}\\
       & & & & 2,000, 3,000, 4,000\} & \\
      & & & & & \\
      Indonesian & Austronesian & Agglutinative  & 3,500 & \{500, 1,000, 1,500\} & \{50, 100, 500\} \\
       & & & & 2,000, 3,000\} & \\
 \hline
    \end{tabular}
    \caption{Descriptive statistics for the languages in our experiments. Data set sizes apply to both sampling strategies (with and without replacement). New test set sizes (sampled without replacement) refer to the different sizes of constructed test sets described in Section~\ref{evaluate2}.}
    \label{tab:languages}
\end{table*}

\section{Meet the Models}

We compared three categories of models: sequence-to-sequence (Seq2seq)~\citep{bahdanau2015neural}, conditional random field (CRF)~\citep{lafferty2001conditional}, and Morfessor~\citep{creutz2002unsupervised}.

\textbf{Seq2seq} The first guest in our model suite is a character-based Seq2seq recurrent neural network (RNN)~\citep{elman1990finding}. 
Previous work~\citep{kann-etal-2018-fortification,liu-etal-2021-morphological} has demonstrated that this model is able to do well for polysynthetic indigenous languages even with very limited amount of data.
Consider the English word \textit{papers} as an illustration.
With this word as input, the task of the Seq2seq model is to perform:

\begin{center}
\textit{papers} $\rightarrow$ \textit{paper + s}.
\end{center}

As we are interested in individual models, the training setups are constant for the data of all languages.
We adopted an attention-based encoder-decoder~\citep{bahdanau2015neural} where the encoder consists of a bidirectional gated recurrent unit (GRU)~\citep{cho-etal-2014-learning}, and the decoder is composed of a unidirectional GRU.
The encoder and the decoder both have two hidden layers with 100 hidden states in each layer.
All embeddings have 300 dimensions.
Model training was carried out with OpenNMT~\citep{klein-etal-2017-opennmt}, using ADADELTA~\citep{zeiler2012adadelta} and a batch size of 16.

\textbf{Order-$k$ CRF} The second guest is the order-$k$ CRF (hereafter $k$-CRF)~\citep{lafferty2001conditional,ruokolainen-etal-2013-supervised}, a type of log-linear discriminative model that treats morphological segmentation as an explicit sequence tagging task.
Given a character $w_{t}$ within a word $w$, where $t$ is the index of the character, CRF gradually predicts the label $y_{t}$ of the character from a designed feature set $x_{t}$ that is composed of local (sub-)strings.

In detail, $k$-CRF takes into consideration the label(s) of $k$ previous characters in a word~\citep{cuong2014conditional,cotterell-etal-2015-labeled}.
This means that when $k=0$ (i.e., $0$-CRF), the label $y_{t}$ is dependent on just the feature set $x_{t}$ of the current character $w_{t}$.
On the other hand, when $k\geq1$, in our settings, the prediction of $y_{t}$ is context-driven and additionally considers a number of previous labels $\{y_{t-k},...,y_{t-1}\}$.

Prior work~\citep{vieira-etal-2016-speed} (see also~\citet{mueller-etal-2013-efficient}) has claimed that, with regards to choosing a value of $k$, 2-CRF usually yields \textit{good} results; increasing $k$ from 0 to 1 leads to large improvements, while increasing from 1 to 2 results in a significant yet weaker boost in model performance.
With that in mind, we investigated five different values of $k$ ($k = \{0, 1, 2, 3, 4\}$).

In training the CRFs, the feature set for every character in a word is constructed as follows. 
Each word is first appended with a start (\textit{$\langle w \rangle$}) and an end (\textit{$\langle /w \rangle$}) symbol.
Then every position \textit{t} of the word is assigned a label and a feature set.
The feature set consists of the substring(s) occurring both on the left and on the right side of the current position up to a maximum length, $\delta$.
As a demonstration, take the same English word above (\textit{papers}) as an example.
For the third character \textit{p} in the word, with $\delta$ having a value of 3, the set of substrings on the left and right side would be, respectively, \{a, pa, $\langle w \rangle$pa\} and \{p, pe, per\}.
The concatenation of these two sets would be the full feature set of the character \textit{p}.
In our experiments, we set $\delta$ to have a constant value of 4 across all languages.

Depending on the position index \textit{t} and the morpheme that this index occurs in,
each character can have one of six labels: \{\textit{START} (start of a word), \textit{B} (beginning of a multi-character morpheme), \textit{M} (middle of a multi-character morpheme), \textit{E} (end of a multi-character morpheme), \textit{S} (single-character morpheme), \textit{END} (end of a word)\}. 
Therefore, \textit{papers} 
has the following segmentation label representation, and the training objective of the CRF model is to learn to correctly label each character given its corresponding feature set.

\begin{tabular}{lllllllllll}
    $\langle w \rangle$ & p & a & p & e & r & s & $\langle /w \rangle$  \\
    START & B & M & M & M & E & S & END
\end{tabular}

In our study, all linear-chain CRFs were implemented with CRFsuite~\citep{CRFsuite} with modifications. 
Model parameters were estimated with L-BFGS~\citep{liu1989limited} and $L_{2}$ regularization.
Again, the training setups were kept the same for all languages.

\textbf{Supervised Morfessor} The last participating model is the supervised variant of Morfessor~\citep{creutz2002unsupervised,rissanen1998stochastic} which uses algorithms similar to those of the semi-supervised variant~\citep{kohonen2010semi}, except that given our crosslinguistic low-resource scenarios, we did not resort to any additional unlabeled data.
For the data of every language, all Morfessor models were trained with the default parameters.

In the remainder of this paper, to present clear narratives, in cases where necessary, we use the term \textit{model alternative(s)} to refer to any of the seven alternatives: Morfessor, Seq2seq, 0-CRF, 1-CRF, 2-CRF, 3-CRF and 4-CRF.

\section{Experiments}

\subsection{Generalizability across data sets}
\label{evaluate1}

Our first question concerns the generalizability of results from the first data set across other data sets of the same size.
For data set construction, we took an approach similar to the synthetic method from~\citet{berg-kirkpatrick-etal-2012-empirical}.
Given the initial data for each language, we first calculated the number of unique word types; after comparing this number across languages, we decided on a range of data set sizes that are small in general (Table~\ref{tab:languages}), then performed random sampling to construct data sets accordingly.

The sampling process is as follows, using Russian as an example. 
For each data set size (e.g., 500), we randomly sampled words with replacement to build one data set of this size; this data set was designated the \emph{first data set}; and then another 49 data sets were sampled in the same way.
We repeated a similar procedure using sampling without replacement.
Each sampled data set was randomly assigned to training/test sets at a 3:2 ratio, five times.
In what follows, we use the term \textit{experimental setting} to refer to each unique combination of a specific data set size and a sampling strategy.

Compared to~\citet{berg-kirkpatrick-etal-2012-empirical}, our data set constructions differ in three aspects. 
First, our sampling strategies include both with and without replacement in order to simulate realistic settings (e.g., real-world endangered language documentation activities) where the training and the test sets may or may not have overlapping items.
Second, in each augmented setting with a specific sampling strategy and a data set size, a total of 250 models were trained given each model alternative (5 random splits $*$ 50 data sets); this contrasts with the number of models for different tasks in~\citet{berg-kirkpatrick-etal-2012-empirical}, ranging from 20 for word alignment to 150 for machine translation.
Third, we aim to build data sets of very small sizes to suit our low-resources scenarios.

For every model trained on each random split, five metrics were computed to evaluate its performance on the test set: full form accuracy, morpheme precision, recall, and F1~\citep{cotterell-etal-2016-morphological-segmentation,van-den-bosch-daelemans-1999-memory}, and average Levenshtein distance~\citep{levenshtein1966binary}.
The average of every metric across the five random splits of each data set was then computed.
For each experimental setting, given each metric, we measured the consistency of model performance from the first data set as follows:

\begin{enumerate}[(1)]
    \item the proportion of times the best model alternative based on the first data set is the best across the 50 data sets;
    \item the proportion of times the model ranking of the first data set holds across the 50 data sets.
\end{enumerate}

If a particular model $A$ is indeed better than alternative $B$ based on results from the first data set, we would expect to see that hold across the 50 data sets.
In addition, the model ranking drawn from the first data set would be the same for the other data sets as well.

While we examined the results of all metrics, for the sake of brevity, we focus on F1 scores when presenting results (Section~\ref{results}).\footnote{Code and full results are in quarantine at \url{https://github.com/zoeyliu18/orange_chicken}.}


\begin{figure*}[h!]
     \centering
\begin{subfigure}{.36\textwidth}
  \includegraphics[width=\linewidth,height=1.8in]{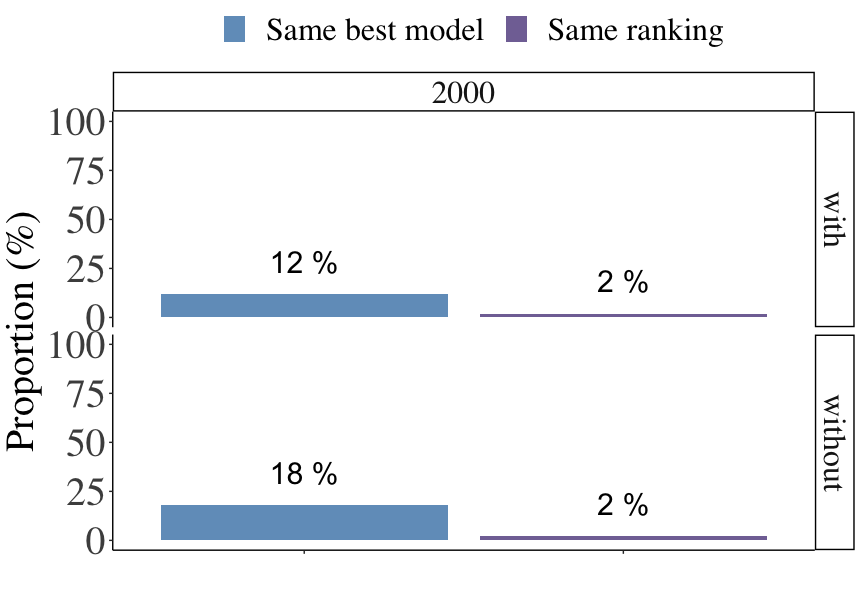}
\caption{Generalizability of results from the first data set across the 50 data sets.}
  \label{fig:russian_sample_best_simple}
\end{subfigure}
\begin{subfigure}{.58\textwidth}
  \includegraphics[width=\linewidth,height=2.3in]{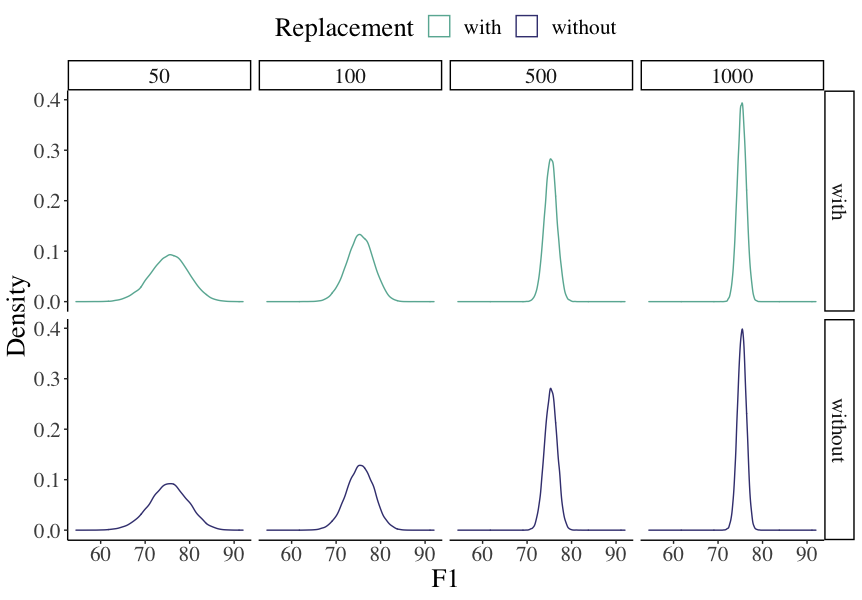}

  \caption{Generalizability across test sets of sizes \{50, 100, 500, 1,000\}}
  \label{fig:russian_test_2000}
\end{subfigure}
         \caption{Generalizability results for the Russian example in Section~\ref{walkthrough}.}
        \label{fig:example}
\end{figure*}

\begin{table*}[h!]
\scriptsize
    \centering
    \begin{tabular}{c|c|c|c|c|c|c}
    \hline
   \textbf{Sampling} &   \textbf{Model} & \textbf{Avg. F1 for} & \textbf{Avg. F1 across} & \textbf{F1 range across} & \textbf{F1 std. across} & \textbf{\% of times} \\ 
   & & \textbf{the first data set} & \textbf{the 50 data sets} & \textbf{the 50 data sets} &  \textbf{the 50 data sets} & \textbf{as the best model} \\ \hline
with & Morfessor &  34.61 &  36.52 & (34.61, 38.36) & 0.76 & 0 \\
 replacement & & & & & & \\
& 0-CRF &  58.75 & 59.59 & (57.83, 62.65) & 0.87 & 0 \\
& & &  & & & \\
& 1-CRF & 74.20 & 75.38 & (73.93, 76.66) & 0.65 & 14 \\
& & & & & & \\
& 2-CRF & 74.02 & 75.46 & (74.02, 76.68) & 0.65 & 22 \\
& & & & & & \\
& 3-CRF & 73.71 & 75.48 & (73.71, 76.52) & 0.67 & 14 \\
& & & & & & \\
& 4-CRF & 74.06 & \textbf{75.54} & (74.06, 76.73) & 0.68 & \textbf{38} \\
& & & & & & \\
& Seq2seq & \textbf{74.26} & 74.86 & (73.48, 76.45) & 0.85 & 12 \\
& & & & & & \\\hline
& & & & & & \\
without & Morfessor & 34.16 & 36.04 &  (34.16, 38.17) & 0.88 & 0 \\
 replacement & & & & & & \\
& 0-CRF & 59.10 & 59.41 &  (57.15, 60.95) & 0.96 & 0 \\
& & & & & & \\
& 1-CRF & \textbf{74.62} & 75.08 & (73.72, 76.72) & 0.72 & 18 \\
& & & & & & \\
& 2-CRF & 74.53 & 75.15 & (73.75, 76.89) & 0.81 & 16 \\
& & & & & & \\
& 3-CRF & 74.60 & 75.18 & (73.77, 76.76) & 0.79 & 18 \\
& & & &  & &\\
& 4-CRF & 74.48 & \textbf{75.21} & (73.79, 77.14) & 0.83 & \textbf{38} \\
& & & & & & \\
& Seq2seq & 74.00 & 74.39 & (72.54, 76.99) & 0.89 & 10 \\

 \hline
    \end{tabular}
    \caption{Summarization statistics of model performance  for the Russian example in Section~\ref{walkthrough}.}
    \label{tab:russian_statistics}
\end{table*}

\subsection{Generalizability across new test sets}
\label{evaluate2}

In addition to evaluating model performance across the originally sampled data sets of the same sizes, we also investigated the generalizability of the best model alternatives from Section~\ref{evaluate1} when facing new unseen test sets.
Taking into consideration the type counts of the initial data for every language and the sizes of their sampled data sets in our experiments, a range of test set sizes that would be applicable to all data set sizes was decided, shown in Table~\ref{tab:languages}. (Note that no new test sets were created for Yorem Nokki and Nahuatl, since it would be feasible for these two languages only when the data set size is 500.)
Then, for every data set, we selected all unique words from the initial data that did not overlap with those in the data set. 
From these words, given a test set size, we performed random sampling without replacement 100 times.
In other words, each data set had correspondingly 100 new test sets of a particular size.

After constructing new test sets, for each experimental setting, we picked the overall best performing model alternative based on average observations across the 50 data sets from Section~\ref{evaluate1}.
For each data set, this model alternative was trained five times (via 5 random splits).
The trained models were then evaluated with each of the 100 new test sets for the data set and the same five metrics of segmentation performance were computed.
Again, due to space constraints, we focus on F1 in presentations of the results.

\section{Results}
\label{results}

\subsection{A Walk-through}
\label{walkthrough}

To demonstrate the two ways of evaluating model generalizability outlined above, take Russian again as an example, when the data set size is 2,000 and the evaluation metric is F1 score.
For the first data set, when sampled with replacement, the best model is Seq2seq, achieving an average F1 score of 74.26. As shown in Figure~\ref{fig:russian_sample_best_simple}, the proportion of times Seq2seq is the best model across the 50 data sets is 12\%.
The model performance ranking for the first data set is Seq2seq $>$ 1-CRF $>$ 4-CRF $>$ 2-CRF $>$ 3-CRF $>$ 0-CRF $>$ Morfessor, a ranking that holds for 2\% of all 50 data sets.

By comparison, when the first data set was sampled without replacement, the best model is 1-CRF with an average F1 score of 74.62; the proportion of times this model has the best results across the 50 data sets is 18\%.
The best model ranking is 1-CRF $>$ 3-CRF $>$ 2-CRF $>$ 4-CRF $>$ Seq2seq $>$ 0-CRF $>$ Morfessor, and this pattern accounts for 2\% of all model rankings considering the 50 data sets together.

Taking a closer look at the performance of each model for the first data set given each sampling strategy (Table~\ref{tab:russian_statistics}), while the average F1 scores of Morfessor and 0-CRF are consistently the lowest, the performance of the other models is quite similar; the difference of F1 scores between each pair of models is less than one.

Now turn to patterns across the 50 data sets as a whole.
It seems that for the Russian example, whether sampling with or without replacement, the best model is instead 4-CRF (Table~\ref{tab:russian_statistics}), which is different from the observations derived from the respective first data set of each sampling method. 
The differences in the proportion of times 4-CRF ranks as the best model and that of other model alternatives are quite distinguishable, although their mean F1 scores are comparable except for Morfessor and 0-CRF. 
With CRF models, it does not appear to be the case that a higher-order model leads to significant improvement in performance.
Comparing more broadly, CRF models are better overall than the Seq2seq models.

In spite of the small differences on average between the better-performing model alternatives, 
every model alternative presents variability in its performance: the score range (i.e., the difference of the highest and the lowest metric score across the 50 data sets of each experimental setting) spans from 2.66 for 2-CRF to 2.97 for Seq2seq.
The difference between the average F1 for the first data set and the highest average F1 score of the other 49 data sets is also noticeable.
When sampling with replacement, the value spans from 2.19 for Seq2seq to 3.90 for 0-CRF;
and when sampling without replacement, the value ranges from 1.85 for 0-CRF to 2.99 for Seq2seq. 

In contrast, in this Russian example, we see a different picture for each model alternative if comparing their results for the first data set to the averages across all the data sets instead; the largest mean F1 difference computed this way is 1.91 when sampling with replacement, and 1.89 for sampling without replacement, both from scores of Morfessor; and these difference values are smaller than the score range from all data sets.

While we focused on F1 scores, we analyzed the other four metrics in the same way. 
Given each metric, the generalizability of the observations from the first data set, regardless of the specific score of each model alternative and their differences, is still highly variable; the most generalizable case, where the best model from the first data set holds across the 50 data sets, is relying on full form accuracy when sampling with replacement (36\%).
When contrasting results from the first data set to the other 49 data sets, there are again noticeable differences between the average score of the former and the highest average score of the latter. 
For instance, when using morpheme precision and sampling with replacement, the difference ranges from 1.98 for Seq2seq to 4.00 for Morfessor; with full form accuracy and sampling without replacement, the difference spans from 2.12 for 0-CRF to 4.78 for Morfessor.

Given each sampling method, we applied the 250 trained models of the best performing model alternative across the 50 data sets (Table~\ref{tab:russian_statistics}) to new test sets of four different sizes.
As presented in Figure~\ref{fig:russian_test_2000}, for this Russian example, 
the distributions of the F1 scores demonstrate variability to different extents for different test set sizes.
When the data was sampled with replacement,
the results are the most variable with a test set size of 50, where the F1 scores span from 57.44 to 90.46;
and the least variable F1 scores are derived when the test set size is 1,000, ranging from 70.62 to 79.36.
On the other hand, the average F1 across the different test sizes is comparable, spanning from 75.30 when the test size is 50 to 75.41 when the test size is 500.
Again, we observed similar results when the data sets were sampled without replacement.

\subsection{An overall look}
\label{overall}

The results for the data sets of all languages were analyzed in exactly the same way as described for the Russian example given in Section~\ref{walkthrough}. When not considering how different the average F1 score of each model alternative is compared to every other alternative within each experimental setting (82 settings in total; 41 for each sampling method), in most of the settings the performance of the model alternatives for the first data set is not consistent across the other data sets; the proportions of cases where the best model alternative of the first data set holds for the other 49 data sets are mostly below 50\%, except for most of the cases for Zulu (data set size $\geqslant$ 1,000) where the proportion values approximate 100\%. 
The observations for model rankings are also quite variable; the proportion of times the model ranking of the first data set stays true across the 50 data sets ranges from 2\% for those in Russian containing 2,000 words sampled with replacement, to 56\% for data sets in Zulu including 4,000 words sampled without replacement.
(The high variability derived this way persists despite of the particular metrics applied.)

\begin{figure*}[h!]
     \centering
\begin{subfigure}{0.8\textwidth}
  \centering
  \includegraphics[width=\linewidth,height=2.5in]{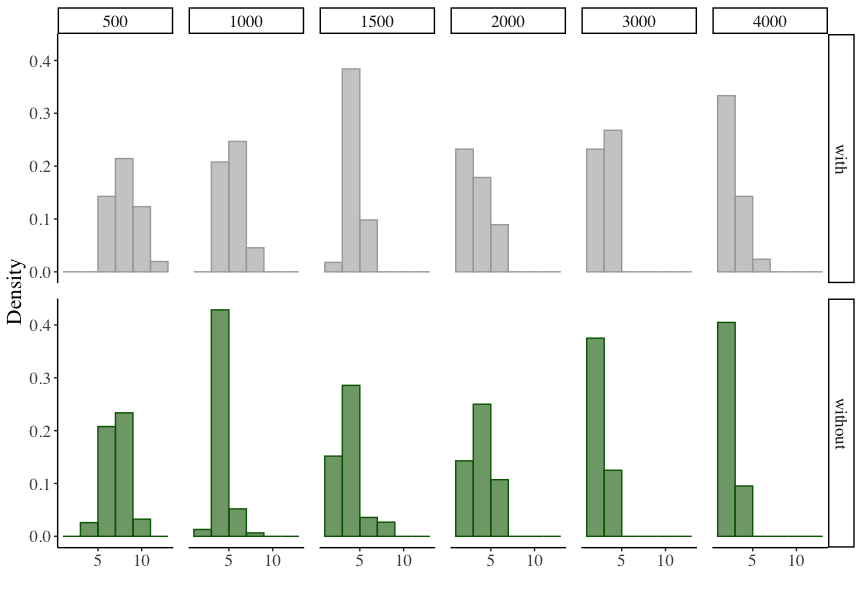}
  \vspace{-0.7cm}
  \caption{Distribution of average F1 score range.}
  \label{fig:data_range}
\end{subfigure}
\vspace{-0.8cm}
\begin{subfigure}{0.8\textwidth}
  \centering
  \includegraphics[width=\linewidth,height=2.5in]{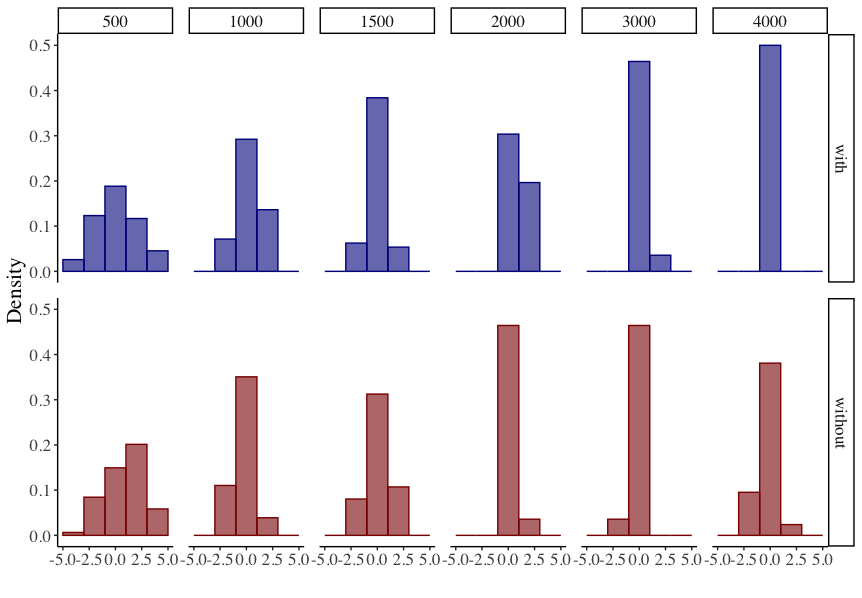}
  \vspace{-0.7cm}
  \caption{Distribution of the difference between the average F1 score of the first data set and that across all 50 data sets.}
  \label{fig:ave_diff}
\end{subfigure}
\vspace{0.8cm}
\caption{Some distribution statistics of the average F1 score of all models across all experimental settings.}
\label{fig:overall_summarization}
\end{figure*}

Considering all the experimental settings, when sampling with replacement, the two best performing models for the first data sets are 4-CRF (13 / 41 = 31.71\%) and Seq2seq (36.59\%); 2-CRF (26.83\%) and 4-CRF (36.59\%) are more favourable if sampling without replacement.
The differences between the best model alternatives are in general small; the largest difference score between the top and second best models is 1.89 for data sets in Zulu with 1,500 words sampled with replacement.
Across the 50 data sets within each setting, overall higher-order CRFs are the best performing models.

Similar findings of performance by different model alternatives were observed for other metrics as well, except for average Levenshtein distance where CRF models are consistently better than Seq2seq, even though the score differences are mostly small. 
This is within expectation given that our task is surface segmentation, and Seq2seq models are not necessarily always able to ``copy" every character from the input sequence to the output.
Previous work has tried to alleviate this problem via multi-task learning~\citep{kann-etal-2018-fortification,liu-etal-2021-morphological}, where, in addition to the task of morphological segmentation, the model is also asked to learn to output sequences that are exactly the same as the input. We opted not to explore multi-task learning here in order to have more experimental control for data across languages.

\begin{figure*}[h!]
\begin{subfigure}{\textwidth}
  \centering
  \includegraphics[width=0.68\linewidth,height=0.245in]{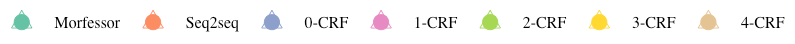}
  \captionsetup{labelformat=empty}
  \label{fig:legend}
\end{subfigure}
\\
\begin{subfigure}{0.33\textwidth}
  \centering
  \includegraphics[width=\linewidth,height=1.55in]{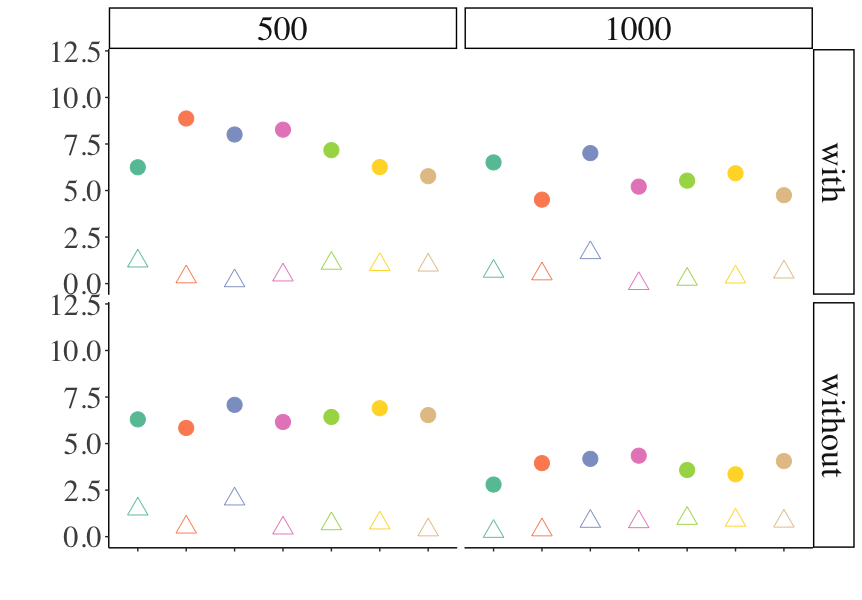}
  \vspace{-1cm}
  \caption{Yorem Nokki}
  \captionsetup{labelformat=empty}
  \label{fig:mayo}
\end{subfigure}
\begin{subfigure}{.33\textwidth}
  \includegraphics[width=\linewidth,height=1.55in]{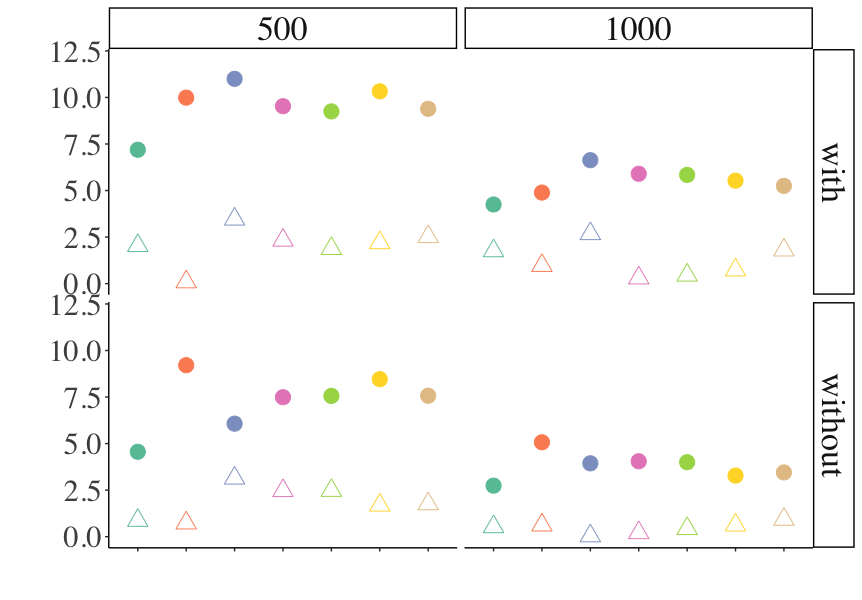}
  \vspace{-1cm}
  \caption{Nahuatl}
  \label{fig:nahuatl}
\end{subfigure}
\begin{subfigure}{.33\textwidth}
  \includegraphics[width=\linewidth,height=1.55in]{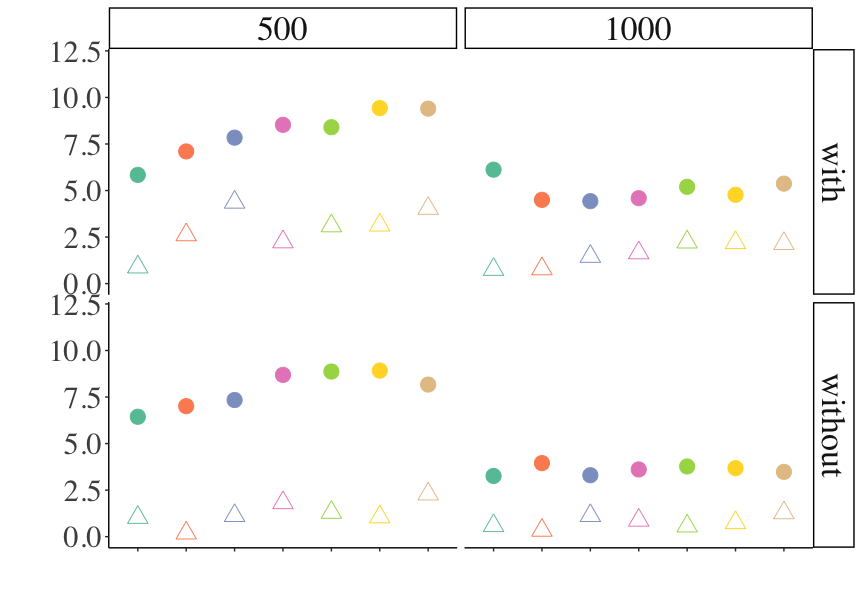}
  \vspace{-1cm}
  \caption{Wixarika}
  \label{fig:wixarika}
\end{subfigure}
\begin{subfigure}{0.4\textwidth}
  \includegraphics[width=\linewidth,height=1.55in]{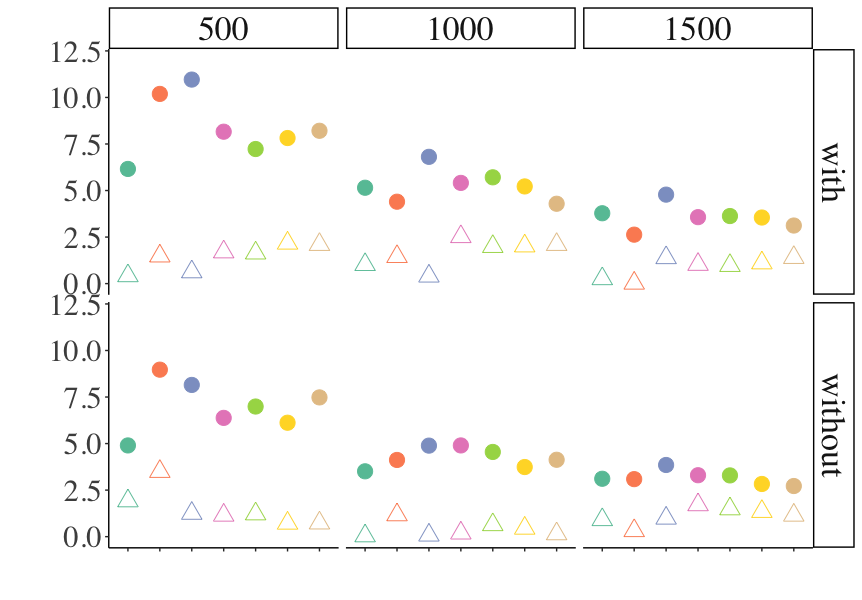}
  \vspace{-1cm}
  \caption{English}
  \label{fig:english}
\end{subfigure}
\begin{subfigure}{.4\textwidth}
  \includegraphics[width=\linewidth,height=1.55in]{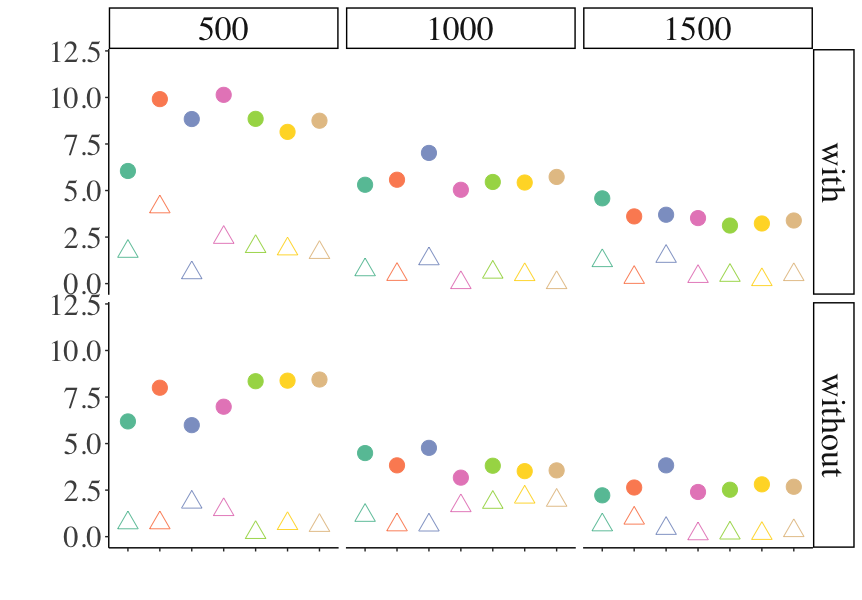}
  \vspace{-1cm}
  \caption{German}
  \label{fig:german}
\end{subfigure}
\begin{subfigure}{.5\textwidth}
  \includegraphics[width=\linewidth,height=1.55in]{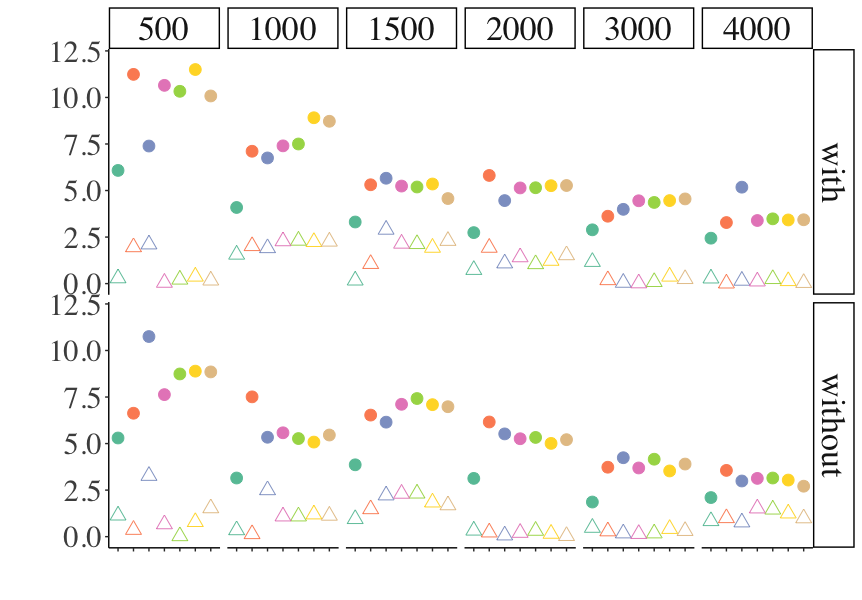}
  \vspace{-1cm}
  \caption{Persian}
  \label{fig:persian}
\end{subfigure}
\begin{subfigure}{.5\textwidth}
  \includegraphics[width=\linewidth,height=1.55in]{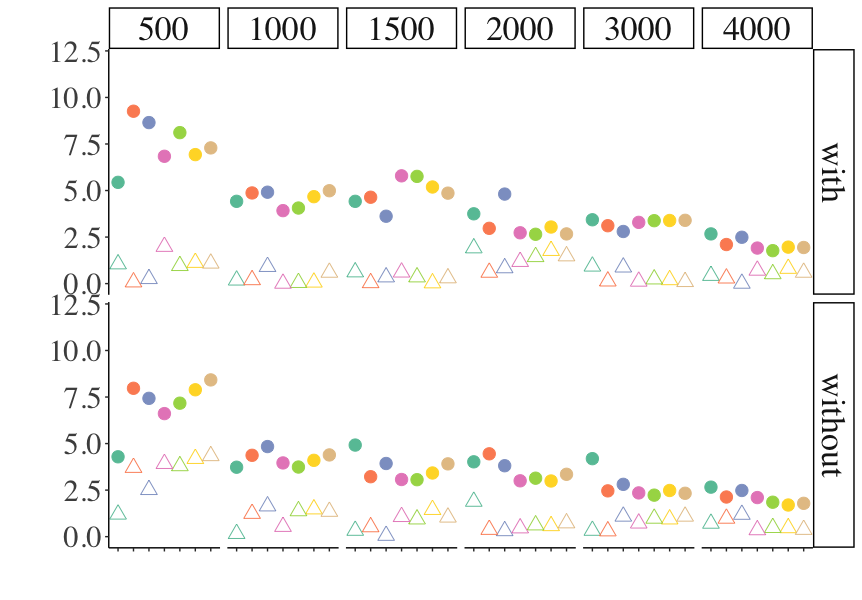}
  \vspace{-1cm}
  \caption{Russian}
  \label{fig:russian}
\end{subfigure}
\begin{subfigure}{0.4\textwidth}
  \includegraphics[width=\linewidth,height=1.55in]{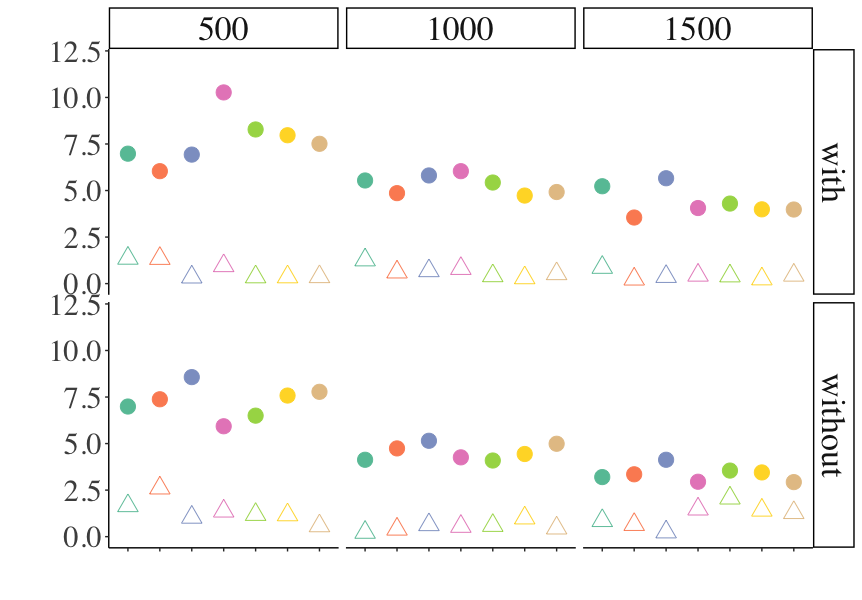}
  \vspace{-1cm}
  \caption{Turkish}
  \label{fig:turkish}
\end{subfigure}
\begin{subfigure}{.4\textwidth}
  \includegraphics[width=\linewidth,height=1.55in]{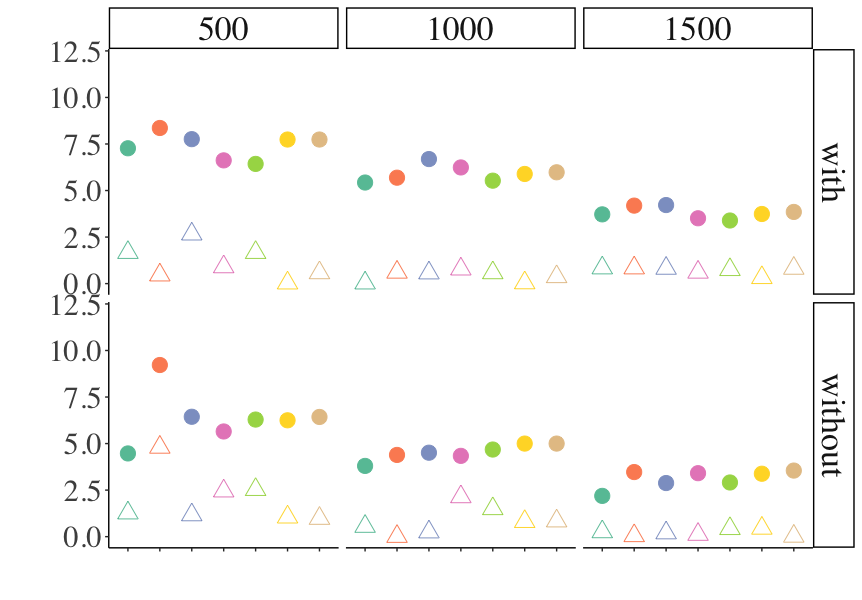}
  \vspace{-1cm}
  \caption{Finnish}
  \label{fig:finnish}
\end{subfigure}
\begin{subfigure}{.5\textwidth}
  \includegraphics[width=\linewidth,height=1.55in]{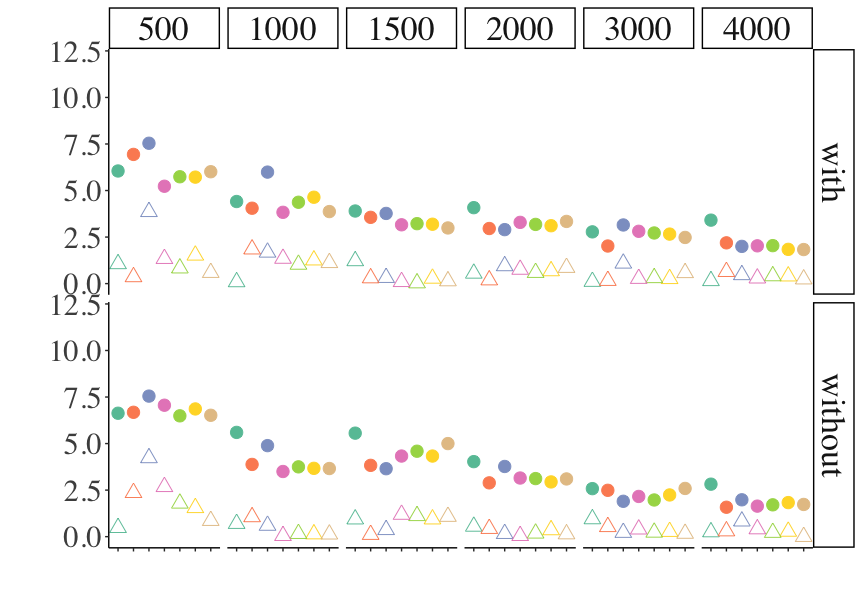}
  \vspace{-1cm}
  \caption{Zulu}
  \label{fig:zulu}
\end{subfigure}
\hspace{0.7cm}
\begin{subfigure}{.45\textwidth}
  \includegraphics[width=\linewidth,height=1.55in]{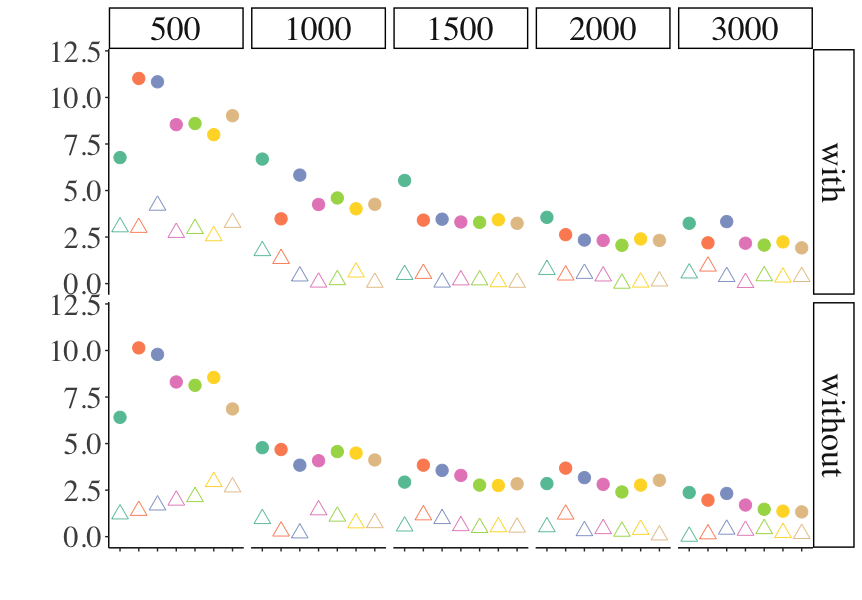}
  \vspace{-1cm}
  \caption{Indonesian}
  \label{fig:indonesian}
\end{subfigure}
\caption{Some summary statistics for the performance of all models across all experimental settings. Within each setting, \textit{circle} represents the average F1 score range; variability exists in the score range regardless of  the model alternative applied or the data set size, though the range appears to be larger when data set sizes are small. By comparison, \textit{triangle} represents the difference between the average F1 score of the first data set and that across all 50 data sets; the values of the differences seem to be more noticeable when data set sizes are small, yet these differences are in general smaller than the score range. }
\label{fig:all}
\end{figure*}

When analyzing the performance range of the models, we see variability as well. 
Figure~\ref{fig:data_range} presents the distribution of F1 score ranges aggregating the results for all the models across all experimental settings; the score ranges of individual model alternatives are additionally demonstrated in Figure~\ref{fig:all}.
Larger score ranges exist especially when data sets have a size of 500, irrespective of the specific model (whether it is non-neural or neural). For example, 3-CRF has a score range of 11.50 for data in Persian sampled with replacement and the range 

\clearpage

\clearpage

\noindent of Seq2seq in this setting is 11.24. As the data set sizes increase, the score ranges become smaller.

Across all experimental settings, we also analyzed the difference between the average F1 score of the first data set and that across all 50 data sets (Figure~\ref{fig:ave_diff}). 
While in most of the cases the differences for the same model alternative are very small, the values of the differences appear to be more distinguishable when the data set size is small.
On the other hand, as shown in Figure~\ref{fig:all}, these differences  are by comparison (much) smaller than the average F1 score range (the \textit{triangles} are consistently below the \textit{circles}), an observation that also aligns with what we have seen from the Russian example.

Again, similar results hold for the other four evaluation metrics as well; different model alternatives demonstrate large score range across all experimental settings regardless of the particular metric.
As an illustration, we performed linear regression modeling analyzing the relationships between other evaluation metrics and the average F1 score. Take morpheme recall as an example. Given all the data of a language, the regression model predicts the score range of average morpheme recall across data sets in an experimental setting as a function of the score range of average F1 in the same experimental setting. The same procedure was carried out for each of the other metrics as well.
Significant effects were found ($p < 0.01$ or $p < 0.001$) for the score range of average F1, indicating strong comparability of model performance in terms of different evaluation metrics.

In addition, besides average F1, there are also noticeable differences between a given metric score for the first data set and that across all 50 data sets within each experimental setting, yet these differences are also smaller than the score range. Our regression models predicting these differences as a function of the differences of average F1 score also show significant effects for the latter, again lending support to the comparability between these metrics when characterizing model evaluation in our study here.

When we applied the trained models of the overall best model alternative to new test sets, all settings demonstrate a great amount of variability that seems to be dependent on both the size of the data set used to train the models and that of the new test sets.
The extent of variation does not appear to be constrained by the sampling strategy.
The difference in the resulting F1 scores ranges from around 6.87, such as the case with Indonesian when the data set size is 3,000 sampled without replacement and the test set size is 500, to approximately 48.59, as is the case for Persian when the data set size is 500 sampled with replacement and the test set size is 50. 
That being said, as in the Russian example, the average results among the different test sizes for each setting of every language are also comparable to each other.\footnote{In addition to testing the best model alternatives, for every experimental setting of Persian, Russian and Zulu, the three languages that have the largest range of data set sizes and new test set sizes in our study, we also experimented with the second or third best model alternatives based on their average performance and computing power. The observations of these alternatives are comparable to those of the best models.}

\begin{table*}[h!]
\scriptsize
    \centering
    \begin{tabular}{c|c|c|c|c|c}
    \hline
   \textbf{Language} &   \textbf{Word overlap} & \textbf{Morpheme overlap} & \textbf{Ratio of Avg. $N$}  & \textbf{Distance between distributions} & \textbf{Ratio of Avg.} \\ 
   & & & \textbf{of morphemes} & \textbf{of $N$ of morphemes} & \textbf{morpheme length} \\\hline
Yorem  Nokki  & \textbf{11.68**} & 17.10 & \textbf{13.64**} & -7.19 & 4.81 \\
  
      & & & & & \\
      Nahuatl  & \textbf{13.15**} & \textbf{49.56***} & \textbf{14.22***} & -3.16 & -1.15 \\
 
      & & & & & \\
      Wixarika  & \textbf{21.57***} & \textbf{63.69***} & -2.58 & -0.70 & \textbf{-11.06**} \\
  
      & & & & \\
      English  & \textbf{9.97**} & \textbf{50.35***} & \textbf{26.01***} & \textbf{-6.78*} & \textbf{8.44*} \\
   
       & & & & \\
      German  & \textbf{10.03**} & \textbf{61.09***} & \textbf{21.44***} & 2.29 & 3.25 \\
 
       & & & & \\
      Persian  & \textbf{26.90***} & \textbf{21.56***} & \textbf{26.15***} & \textbf{-3.82*} & -3.09  \\

       & & & & \\
      Russian  & 3.38 & \textbf{69.49***} & \textbf{11.96***} & \textbf{-2.88**} & 3.19 \\

       & & & & \\
      Turkish  & \textbf{15.88***} & \textbf{44.31***} & 1.37 & -0.73 & 0.30 \\

   & & & & \\
      Finnish  & \textbf{9.47**} & \textbf{60.58***} & \textbf{10.49**} & -1.95 & -3.90 \\

      & & & & \\
      Zulu  & \textbf{15.48***} & \textbf{79.07***} & \textbf{11.34***} & \textbf{-4.14***} & \textbf{4.11} \\

      & & & & \\
      Indonesian  & \textbf{8.12*} & \textbf{25.53***} & \textbf{19.55***} & \textbf{-6.46**} & \textbf{7.64*} \\

 \hline
    \end{tabular}
    \caption{Regression coefficients for random splits of data sets in all languages; a positive coefficient value corresponds to higher metric scores; the numbers in bold indicate significant effects; the number of * suggests significance level: * $p < 0.05$, ** $p < 0.01$, *** $p < 0.001$.}
    \label{tab:regression}
\end{table*}

\subsection{Regression analysis}
\label{variable}

One question arises from the aforementioned findings: why is there variable performance for each model alternative across all the experimental settings of each language?
With the observations in our study thus far, it seems that the results are dependent on the data set size or test set size.
But is that really the case, or is sample size simply a confounding factor?

To address the questions above, we studied several \textit{data characteristics} and how they affect model performance. It is important to note that, as discussed in Section~\ref{introduction}, we do not wish to claim these features are representative of the full language profile (e.g., how morphologically complex the language is as a whole).
For instance, one might expect that on average, languages that are agglutinative or polysynthetic have larger numbers of morphemes per word when compared to fusional languages. 
While that might be true for cases with ample amounts of data, the same assumption does not always seem to be supported by the data sets in our experiments.
For each experimental setting, the average number of morphemes per word across the 50 data sets of Indonesian (classified as agglutinative) is mostly comparable to that of Persian (fusional); on the other hand, in all experimental settings that are applicable, the average number of morphemes per word for data sets of Russian (fusional) is always higher than that for data of the Mexican indigenous languages (polysynthetic).

These patterns resonate again with the our main point, that when data set size is small, the first data set or just one data set might not suffice to reflect the language features overall.
Thus here we consider the characteristics to be specific to the small data sets in our experiments.
For each random split (including a training and a test set) of every data set, we investigated: 
(1) word overlap, the proportion of words in the test set that also occur in the training set (only applicable to sampling with replacement);
(2) morpheme overlap, the proportion of morphemes in the test set that also appear in the training set;
(3) the ratio of the average number of morphemes per word between the training and test sets;
(4) the distance between the distribution of the average number of morphemes per word in the training set and that in the test set; for this feature we used the Wasserstein distance for its ease of computation~\citep{arjovsky2017wasserstein,sogaard-etal-2021-need};
(5) the ratio of the average morpheme length per word between the training and the test sets.

To measure the individual effect of each feature, we used linear regression.
Because of the relatively large number of data sets that we have and our areas of interest, we fit the same regression model to the data of each language rather than combining the data from all languages.
The regression predicts the metric scores of all models for each random split as a function of the five characteristics, described above, of the random split.
Meanwhile we controlled for the roles of four other factors: the model alternatives, the metrics used, sampling methods, and data set size, the latter two of which also had interaction terms with the five characteristics.

Given the components of our regression model, a smaller/larger value of the coefficient for the same factor here does not necessarily mean that this factor has a weaker/stronger role. In other words, the coefficients of the same feature are not comparable across the data for each language. Rather our goal is simply to see whether a feature potentially influences metric scores when other factors are controlled for (e.g., data set size) within the context of the data for every language.

The regression results are presented in Table~\ref{tab:regression}. 
It appears that the features with the most prominent roles are the proportion of morpheme overlap as well as the ratio of the average number of morphemes per word. Word overlap also has significant effects, though this is applicable only to data sets sampled with replacement.
In comparison, data set size does not appear to have significant positive effects in all cases; the exceptions include the data for Yorem Nokki, Nahuatl, Wixarika, and Turkish.
In scenarios where data set size does have an effect, its magnitude is much smaller compared to the characteristics in Table~\ref{tab:regression} that we studied.
Thus, while data set size potentially plays a role in the results, it does not appear to be the sole or even the most important factor. The range of model performance is more dependent on the specific features of, or what is available in, the data sets. For example, larger training and test sets do not necessarily lead to higher morpheme overlap ratio.

We adopted similar approaches to investigate the variability in the results when trained segmentation models were applied to new test sets of different sizes (Section~\ref{evaluate2}).
In these cases, the five characteristics described before were measured taking into account the training set of the segmentation model and each of the new test sets (100 in total for each training set).
In contrast to the previous regression model, since the new test sets were sampled without replacement
and since we are focusing on predictions derived from the best model alternative (but see Section~\ref{overall}), word overlap ratios and the model alternatives were not included in the regression. 
We additionally controlled for the effect of the new test set size.
Again the same regression model was fit to the data of each language.
Comparing the different features, morpheme overlap ratio and the average number of morphemes per word are again the two most pronounced factors in model performance, while the role of test set size is much less pronounced or even negligible.

\subsection{Alternative data splits}
\label{notes}

Our experiments so far involved random splits.
While the heuristic or adversarial splits proposed in~\citet{sogaard-etal-2021-need} are also valuable, the success of these methods requires that the same data \textit{could} be split based on heuristics, or be separated into training/test sets such that their distributions are as divergent as possible.

To illustrate this point for morphological segmentation, we examined the possibility of the data sets being split heuristically and adversarially.
For the former, we relied on the metric of the average number of morphemes per word, based on regression results from Section~\ref{variable}.
Given each data set within every experimental setting, we tried to automatically find a metric threshold so that the words in the data set are able to be separated by this threshold into training/test sets at the same 3:2 ratio, or a similar ratio; words in which the number of morphemes goes beyond this threshold were placed in the test set.
We note that it is important (same as for adversarial splits) that the number of words in the resulting training and test sets follows a similar ratio to that for random splits. 
This is in order to ensure that the size of the training (or the test) set would not be a factor for potential differences in model performance derived from different data split methods.
In most of the data sets across settings, however, such a threshold for the average number of morphemes per word does not seem to exist. The exceptions are Wixarika, where the maximum number of data sets that are splittable this way is 35, for data sets containing 1,000 words sampled without replacement; and certain cases in Finnish, where the maximum number of data sets suitable for heuristic splitting is 11, when the data sets have 500 words sampled with replacement.

For adversarial splitting, 
we split each data set (for five times, as well) via maximization of the Wasserstein distance between the training and the test sets~\citep{sogaard-etal-2021-need}. 
We then calculated the word overlap ratio between the test sets of the adversarial splits and those from the random splits.
Across most of the experimental settings, the average word overlap ratios center around or are lower than 50\%.
This suggests that the training and test sets of adversarial splits are reasonably different from those derived after random splits; in other words, these data sets could be split adversarially.

While it is not the focus in this study to compare different ways of splitting data, we carried out adversarial splits for the data of Persian and Russian, two of the languages with the largest range of data set sizes and new test set sizes.
The models applied were second-order and fourth-order CRFs because of their overall good performance (Section~\ref{results}).
For each experimental setting, the results are still highly variable. 
Compared to splitting data randomly, the average metric scores of adversarial splits are lower (e.g., the mean difference of the average F1 scores between the two data split methods is 17.85 across settings for the data of Zulu, and 6.75 for the data of Russian), and the score ranges as well as standard deviations are higher.
That being said, the results from these two data split methods are, arguably, not directly comparable, since with adversarial splits the test sets are constructed to be as different or \textit{distant} as possible from the training sets.

When applying the trained models to the same new test sets sampled from Section~\ref{evaluate2}, the observations are similar to the descriptions above, except the differences between the two data split methods regarding the mean metric scores, score ranges, and variances are much smaller. The most pronounced average difference in F1 is approximately 7.0 when the experimental settings yield small data set sizes. 
On the other hand, within each setting, despite the model performance variability, the average results of the different test sizes are again comparable.

\section{Discussion and Conclusion}

Using morphological segmentation as the test case, we compared three broad classes of models with different parameterizations in crosslinguistic low-resource scenarios.
Leveraging data from 11 languages across six language families, our results demonstrate that the best models and model rankings for the first data sets do not generalize well to other data sets of the same size, though the numerical differences in the results of the better-performing model alternatives are small.
In particular, within the same experimental setting, there are noticeable discrepancies in model predictions for the first data set compared to the averages across all data sets; and the performance of each model alternative presents different yet significant score ranges and variances.
When examining trained models on new test sets, considerable variability exists in the results.
The patterns described above speak to our concerns raised at the beginning, namely, that when facing a limited amount of data, model evaluation gathered from the first data set -- or essentially any one data set -- could fail to hold in light of new or unseen data.

To remedy the observed inconsistencies in model performance, we propose that future work should consider utilizing random sampling from initial data for more realistic estimates.
We draw support for this proposal from two patterns in our study.
First, in each experimental setting, the difference between the average F1 score of the first data set and that across all 50 data sets is in general smaller than the range of the scores.
Second, the average metric scores across unseen test sets of varying sizes in every setting are comparable to each other; this holds for models trained from random splits and those trained using adversarial splits.

Therefore, depending on the initial amount of data, it is important to construct data sets and new test sets of different sizes, then evaluate models accordingly.
Even if the size of the initial data set is small, as is the case with most endangered languages, it is worthwhile to sample with replacement in order to better understand model performance. 
Based on these observations, it would be interesting to analyze the potential factors that lead to the different degrees of generalization, which could in turn provide guidance on what should be included or sampled in the training sets in the first place.
Thorough comparisons of different models should be encouraged even as the amount of data grows, particularly if the experimental settings are still situated within low-resources scenarios for the task at hand~\citep{hedderich2020survey}.

Lastly, while we did not perform adversarial splits for all experimental settings here, for the cases that we have investigated, model performance from adversarial splits also yields high variability both across data sets of the same size and in new test sets.
Though our criteria for heuristic splits were not applicable with the data explored, we would like to point out that for future endeavors, it would be necessary to at least check the applicability of different data splits, as we did in Section~\ref{notes}.
As long as the data set possesses the properties that allow it to be divided by different split methods, it is worthwhile to further explore the influence of these splits, coupled with random sampling.

\section*{Acknowledgements}

We are grateful to the reviewers and the action editor for their insightful feedback.
This material is based upon work supported 
by the National Science Foundation under
 Grant \#2127309 to the Computing Research 
Association for the CIFellows Project, 
and Grant \#1761562.
Any opinions, findings, and conclusions or recommendations expressed in this material are those of the author(s) and do not necessarily reflect the views of the National Science Foundation nor the Computing Research Association.


\bibliography{tacl2018,custom}
\bibliographystyle{acl_natbib}

\end{document}